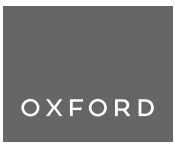

Data and text mining

# Sequence tagging for biomedical extractive question answering

Wonjin Yoon [1,†], Richard Jackson[2], Aron Lagerberg[3,]* and Jaewoo Kang [1,4,]*

[1]Department of Computer Science and Engineering, Korea University, Seoul 02841, South Korea, [2]AstraZeneca UK, Cambridge CB2 0AA, UK, [3]AstraZeneca SE, 43150 Mölndal, Sweden and [4]AIGEN Sciences Inc., Seoul 04778, South Korea

*To whom correspondence should be addressed.
[†]This work was done while Wonjin Yoon worked under the Research Collaboration project at AstraZeneca.
Associate Editor: Jonathan Wren

## Abstract

**Motivation:** Current studies in extractive question answering (EQA) have modeled the single-span extraction setting, where a single answer span is a label to predict for a given question-passage pair. This setting is natural for general domain EQA as the majority of the questions in the general domain can be answered with a single span. Following general domain EQA models, current biomedical EQA (BioEQA) models utilize the single-span extraction setting with post-processing steps.
**Results:** In this article, we investigate the question distribution across the general and biomedical domains and discover biomedical questions are more likely to require list-type answers (multiple answers) than factoid-type answers (single answer). This necessitates the models capable of producing multiple answers for a question. Based on this preliminary study, we propose a sequence tagging approach for BioEQA, which is a multi-span extraction setting. Our approach directly tackles questions with a variable number of phrases as their answer and can learn to decide the number of answers for a question from training data. Our experimental results on the BioASQ 7b and 8b list-type questions outperformed the best-performing existing models without requiring post-processing steps.
**Availability and implementation:** Source codes and resources are freely available for download at https://github.com/dmis-lab/SeqTagQA.
**Contact:** aronlagerberg@gmail.com or kangj@korea.ac.kr
**Supplementary information:** Supplementary data are available at *Bioinformatics* online.

## 1 Introduction

Extractive question answering (EQA) is the process of finding answers to questions from given passages. Biomedical Question Answering (BioQA) is a branch of QA where both the subject of the question and the passage has a biomedical context (Fig. 1). EQA has the potential to assist in the management of the deluge of textual data arising from scientific research, as it can be deployed upon vast datasets of scientific literature that are impractical for individuals or teams of scientists to digest. The use of pretrained transformer architectures such as BERT, GPT or RoBERTa have yielded rapid improvements in the general domain EQA datasets, even outperforming humans in SQuAD datasets (Rajpurkar *et al.*, 2016, 2018). Accordingly, such models have also shown promise in the BioQA domain. Yoon *et al.* (2019b) utilized BioBERT (Lee *et al.*, 2020), with an expanded span detection configuration to solve biomedical questions. This architecture produced the best performance for both factoid and list question on the BioASQ 7b dataset.

Despite recently increased interests in the field of BioQA, research on the characteristics of biomedical questions and corresponding passages is rare. As Friedman *et al.* (2002) and Mollá and Vicedo (2007) suggested, semantics and syntactic structures of specialized domains, such as biomedical literature, are different from general domain corpora. The existence of such differences implies that rapid advancements in general domain QA may not transfer well to BioQA. We believe that a deeper understanding of these differences will guide the development of methods in BioQA. As an effort to understand the difference, we conduct a preliminary study on the characteristics of biomedical questions and find that the proportions of *list questions* (i.e. questions that have multiple answers; a list of answers) are more abundant in biomedical questions than in general domain questions.

The single-span extraction setting, or 'start-end span prediction' setting, is common in existing general domain EQA and Biomedical EQA (BioEQA) models (Jeong *et al.*, 2020; Wiese *et al.*, 2017; Yoon *et al.*, 2019b). Single-span prediction setting is widely used as it is [Factoid] **Question** : Which is the most common gene signature in Rheumatoid Arthritis patients?

**Passage**: … Patients with systemic lupus erythematosus, Sjögren's syndrome, dermatomyositis, psoriasis, and a fraction of patients with rheumatoid arthritis display a specific expression pattern of interferon-dependent genes in their leukocytes, termed the **interferon signature**. Here, in an … O O O B I O O O … attempt to understand the role of type I interferons in the pathogenesis of autoimmunity, we review the recent …

[List] **Question** : Which are the clinical characteristics of isolated Non-compaction cardiomyopathy?

**Passage**: … The form of presentation was **heart failure** in … O B I O 53% of subjects, **syncope** in 20%, **ventricular arrhythmias** O O O B O O B I in 13% and **stroke** in 7%. Left ventricular end-diastolic … O O O B O …

**Fig. 1.** Examples of two types of BioEQA. Factoid questions require a phrase as an answer while list questions require multiple phrases as their answer. Answers are underlined in the corresponding passage and annotated using BIO tagging scheme

straightforward and effective approach for single answer questions. However, this approach has limitations. For questions where more than one answer spans exist within the given passage, a model trained in the single-span prediction setting will only treat one answer candidate as true label and the others as noise, despite the other answers also being correct. Previous works on list questions require complex post-processing steps to determine the number of predicted answers. As an example, Yoon et al. (2019b) and Jeong et al. (2020) applied a fixed probability threshold strategy which needed to be learned using a validation dataset, and rule-based query processing to detect the number of answers existing in the question. Related to this issue, the BioASQ dataset classifies questions into categories such as factoid-type question (single answer) or list-type question (multiple answer). However, in a real-world setting, such metadata is generally unavailable. Therefore, a more flexible assumption is that all questions will produce a list of answers (even if the list is empty, or only contains a single answer).

In order to alleviate the shortcomings of single-span extraction setting, we propose to reformulate the task of BioEQA as sequence tagging (i.e. a multi-span extraction setting). We empirically verify that the sequence tagging approach is beneficial for answering list-type questions, as models are able to learn to tag multiple answers at once rather than being restricted to extract one answer span at a time. Furthermore, our setting is an end-to-end approach for list-type QA task that the model can learn to output the ideal number of answers for a given question, without using the rather complicated external processes of previous works. Adopting our sequence tagging strategy to BioEQA tasks showed that our model can achieve state-of-the-art (SOTA) performance for list-type questions. Average performance improvements over baseline model are 3.80% and 6.22% for BioASQ 7b and 8b List questions, respectively (F1 score).

## 2 Related work

In this section, we briefly summarize previous works related to the questions in QA datasets and various modeling approaches for QA tasks.

### 2.1 Question answering datasets

A number of general domain QA datasets are available as of today. SQuAD v1 and v2 (Rajpurkar et al., 2016, 2018) are composed of Wikipedia articles and questions generated by crowdsourcing, and become one of the most visited datasets for EQA task. In contrast to the aforementioned studies where the questions are generated by annotators, some works focus on how questions are collected. MS MARCO (Nguyen et al., 2016), DuReader (He et al., 2018) and Natural Questions (NQ) (Kwiatkowski et al., 2019) datasets use search engine queries submitted by users to collect questions as an effort to harmonize the distribution of question to the needs of real user questions; these questions are often referred as 'naturally posed' questions in contrast to questions generated artificially by annotators. Another branch of QA datasets includes yes/no questions (Clark et al., 2019) and cloze style questions (Hermann et al., 2015). However, yes/no and cloze questions are rarely covered by models with EQA settings.

In comparison to the general domain, only a small number of QA datasets are currently publicly available for Biomedical NLP. The QA dataset of the BioASQ competition (Tsatsaronis et al., 2015) is considered as one of the richest source for BioEQA models. The BioASQ dataset offers factoid, list, yes/no and summarization questions along with passages and answers. Besides EQA datasets, a multiple-choice QA dataset focused on Alzheimer's disease (Morante et al., 2013) is released as a pilot task of the Question Answering for Machine Reading Evaluation (QA4MRE). PubMedQA (Jin et al., 2019) created a QA dataset that can be used as yes/no or query-focused summarization. Cloze style QA datasets are also proposed in the domain of BioNLP (Kim et al., 2018; Lamurias et al., 2020; Pappas et al., 2020).

In contrast to the pool of accessible BioQA datasets, naturally posed biomedical questions have rarely been considered as research material. Analogous to google queries for NQ dataset (Kwiatkowski et al., 2019), a one-day log of PubMed queries is available in raw-data format (Herskovic et al., 2007). Yet, so far the PubMed queries have not been refined and analyzed as a source of questions. A group of researchers released a collection of questions, namely Clinical Questions Collection (CQC) Data, which consists of question that are asked by physicians while caring for their patients (D'Alessandro et al., 2004; Ely et al., 1999, 1997). We will discuss naturally posed questions in Section 3.

### 2.2 Question answering models

*QA models in BioNLP* Wiese et al. (2017) applied the FastQA model (Weissenborn et al., 2017) for BioQA and saw a gain in performance. To the best of our knowledge, Wiese et al. (2017) was the first attempt to use a neural network (NN)-based QA model to achieve first place for the EQA problems of the fifth BioASQ challenge Task b—Phase B. BioBERT (Lee et al., 2020) is a BERT model trained on biomedical data and showed a large performance improvement over preceding models for the BioEQA. Yoon et al. (2019b) and Jeong et al. (2020) won the seventh and eighth BioASQ challenges, respectively, using BioBERT as a core building block for the factoid, list and yes/no questions (Nentidis et al., 2020a, b). The aforementioned NN-based models for BioEQA (Jeong et al., 2020; Lee et al., 2020; Wiese et al., 2017; Yoon et al., 2019b) formulated the training objective of their models as a single-span prediction.

*QA and other NLP tasks* A number of researchers have utilized the methods of QA frameworks to solve various NLP tasks. Recently, Li et al. (2020, 2019) exhibited benefits of the QA framework for the Named Entity Recognition (NER) and Relation Extraction (RE) tasks. Li et al. (2020) utilized single-span extraction setting for the NER-task while Li et al. (2019) tackled the RE task by using an NER-like sequence tagging approach, in order to predict multiple phrases from a sample. Both work from Li et al. (2020, 2019) benefitted from pre-trained language models such as BERT. Apart from the recent QA approaches, Yao et al. (2013) used Conditional Random Field (CRF)-based model with the Tree Edit distance feature for QA-task and applied a sequence tagging approach to predict the location of answer tokens in candidate sentences. Segal et al. (2020) utilized a multi-head model approach and tag-based span extraction for question answering task in general domain and reached state-of-the-art performance.

To the best of our knowledge, no preceding works have applied sequence tagging approaches to tackle the BioEQA tasks, nor studied the difference of question distributions between general and biomedical domain.

## 3 Biomedical questions

In this section, we elaborate on our preliminary study on the distribution of biomedical questions. We collected biomedical questions from existing literature that provide *naturally posed* questions asked by medical doctors and PubMed search engine users. The questions are categorized by required answer types: factoid, list, yes/no and summarization. Preliminary findings support our hypothesis that models which can generate one or more responses for a given input will improve the performance on BioEQA tasks.

### 3.1 Types of questions

We first hypothesize that biomedical questions have a tendency to require multiple answers than their general domain counterparts since biomedical concepts tend to involve collections of multiple entities. For instance, diseases often have multiple symptoms, biochemical pathways involve multiple genes and a variety of drugs are often used to treat a given disease. To provide supporting evidence for our assumption, we analyze the characteristics of biomedical questions. Specifically, our objective of the analysis is to empirically reveal the different distributions of naturally posed biomedical questions and general domain questions.

Referring to the **Natural Questions (NQ)** (Kwiatkowski *et al.*, 2019) and **BioASQ** datasets for QA (Tsatsaronis *et al.*, 2015), we created the following question categories:

- *Long answer only*: A question that requires a sentence-length description as an answer. Corresponds to Summary questions of BioASQ.
- *Short answer—single phrase* or *Factoid*: A question that requires a phrase as an answer set. Corresponds to Factoid questions of BioASQ.
- *Short answer—multiple phrases* or *List*: A question that requires more than one phrase as an answer set. Corresponds to List questions of BioASQ.
- *Binary answer* or *Yes/No*: A question that is answerable with either Yes or No. Corresponds to Yes/No questions of BioASQ.
- *Others*: A question that is not answerable with current settings. Corresponds to Bad- and 'No answer'-questions of NQ.

### 3.2 Sources of questions

The sources of questions we have analyzed are the following:

*Clinical questions collection* Clinical Questions Collection (CQC) is a series of datasets (D'Alessandro *et al.*, 2004; Ely *et al.*, 1999, 1997) that are collected by observing healthcare professionals such as primary care physicians, faculty of university hospitals and residents and medical students during clinical sessions (https://www.nlm.nih.gov/databases/download/CQC.html). We select IOWA IC QUESTIONS as our source of questions as it is collected from the most diverse set of physicians among the CQC datasets. The IOWA IC QUESTIONS dataset contains 1,062 questions that were collected by observing 48 physicians for a half-day.

*PubMed queries* We use the PUBMED QUERY LOG DATASET collected by Herskovic *et al.* (2007), which consists of logs of *naturally posed* questions by PubMed users collected over one day. (More details including *query screening algorithm* in Supplementary Appendix.)

Corresponding passages or answers are not provided for both biomedical question datasets as they are not originally built for QA tasks. We use the **Natural Questions (NQ)** dataset as a question dataset in the general domain setting.

### 3.3 Results and meanings of the preliminary study

We classify a set of randomly chosen biomedical questions ($n = 100$; 50 for each biomedical question dataset) manually. For general domain questions, we counted questions using the given answers of the NQ dataset. Figure 2 shows the distribution of naturally posed questions in the general domain (NQ) and biomedical domain (CQC and PubMed).

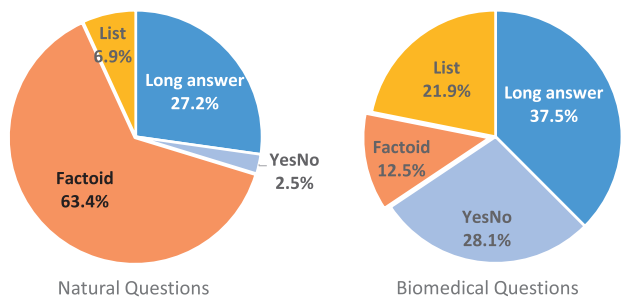

**Fig. 2.** Distribution of naturally posed questions in the general and biomedical domains

Our study shows that the proportion of factoid questions (i.e. questions which can be answered with a single phrase) is lower in biomedical questions than in questions in the general domain. In other words, the number of list-type questions among EQA questions, specifically factoid and list-type questions, is significantly higher in the biomedical context than in the general context. This finding verifies our hypothesis that biomedical questions have a tendency to require multiple answers.

While annotating the types of questions, we observed that factoid questions often have alternative valid answer spans. In most cases, they are varied forms of the originally annotated answer, such as synonyms or abbreviations. The prevalence of synonyms and abbreviations for biomedical entities (Sung *et al.*, 2020), which form most of the answers, makes it natural to shift the conventional approach of BioEQA question and answers being in a one-to-one structure to a one-to-many structure (We will provide access to our question collection online.).

## 4 Sequence tagging for question answering

In this section, we describe our strategy for BioEQA. EQA aims to detect the answer spans in a provided passage given a question. Instead of the conventional single-span extraction setting, or start-end prediction strategy, we applied a sequence tagging framework to the QA task in order to detect a variable number of answer spans.

### 4.1 Contextualized representation for biomedical NLP tasks

Biomedical Language Models (LMs) are language representation models that are pre-trained on biomedical literature such as articles from PubMed and PubMed Central. They are the primary building blocks of multiple BioNLP tasks, such as NER, RE and QA tasks, and provide contextualized representations of the input sequences.

In our experiments, BioBERT (Lee *et al.*, 2020), BlueBERT (Peng *et al.*, 2019) and PubMedBERT (Gu *et al.*, 2021) are used as a biomedical LM for our model. BioBERT is the first Bidirectional Encoder Representations from Transformers (BERT) model pre-trained on the biomedical corpora. Pre-training of BioBERT and BlueBERT is continued from BERT (Devlin *et al.*, 2019) and therefore the vocabulary of BioBERT and BlueBERT is identical to the BERT models. PubMedBERT is pre-trained from scratch with the vocabulary built from the biomedical corpora. BioBERT used the Cased model, whereas BlueBERT and PubMedBERT used the Uncased model: for the latter one, input sentences are lower-cased during the pre-processing steps.

### 4.2 Sequence tagging for question answering

Figure 3 illustrates the overall structure of our model for EQA.

#### 4.2.1 Input sequence

Following BERT and BioBERT, we concatenate the question to the corresponding passage to form an input sequence and use the WordPiece tokenization (Wu *et al.*, 2016) to alleviate the out-of-vocabulary

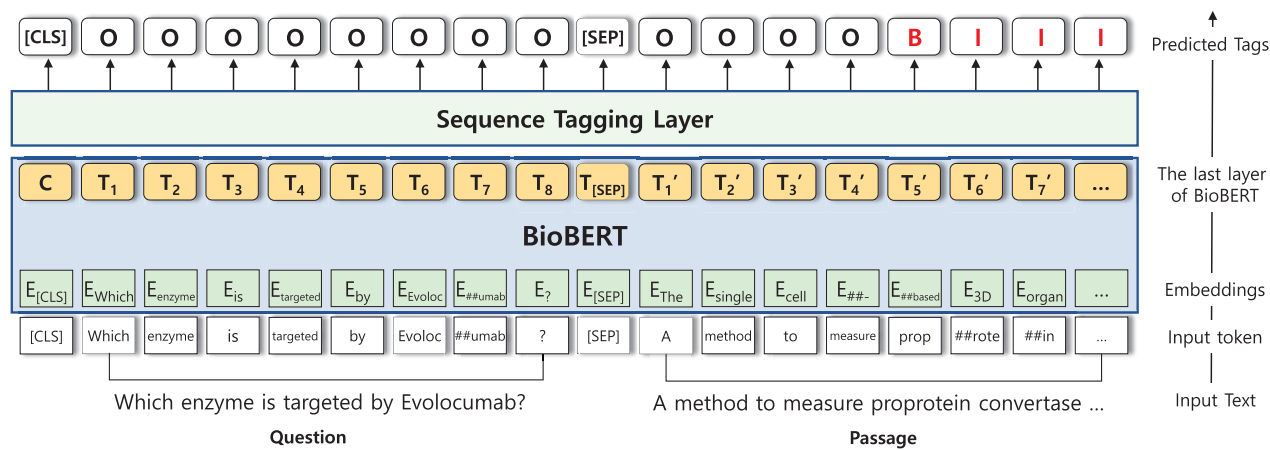

**Fig. 3.** Overview of the BioBERT model performing QA as sequence tagging. A question and passage from a sample forms the input token sequence after tokenization. The input sequence is fed into BioBERT to output the contextualized representations. The final layer of the model is a sequence tagging layer which predicts a tag/label for each token representation

problem. Special tokens [CLS] and [SEP] are added at the beginning of the sequence and between the question and the passage, respectively. The output of the final layer of BioBERT is fed into the sequence tagging layer to produce the final tags.

#### 4.2.2 Sequence tagging layer

We used neural network-based sequence tagging structures as the final layer. Any structure predicting token-level labels can be used for the final layer. In our experiments, we selected a feed-forward network as the sequence tagging layer. In additions to this, the performance of Bidirectional Long Short-Term Memory (BiLSTM) (Hochreiter and Schmidhuber, 1997; Schuster and Paliwal, 1997) and BiLSTM networks with a Conditional Random Field layer (BiLSTM-CRF) (Huang et al., 2015)-based models are evaluated (Section 5.5).

Following NER tasks, we used the BIO tag schema to denote answer spans, where B stands for the 'beginning' of an answer span, I for 'inside' and O for 'outside'.

When calculating loss and gradients, we exclude tags predicted at the positions of special tokens and question tokens: this means that the model will not be penalized for predicting wrong tags for these tokens. Tokens that we ignored predictions are as follow: [SEP], [CLS], tokens that compose questions and *broken* sub-word tokens.

#### 4.2.3 Post-processing

We apply minimal post-processing steps to the output of a tagged sequence. The output of the final layer (i.e. the sequence tagging layer) is a sequence of tokenized words and their corresponding tags. The main purpose of the post-processing step is to *detokenize* the output of the final layer to restore tokenized strings into their original forms, including reconstruction of sub-word tokens and proper removal of whitespace around punctuation marks. Other essential task-specific post-processing steps are taken to fit our output to the official evaluation scripts for the datasets. For example, answer candidates of list questions of BioASQ can be prepared by simple concatenation of predictions of the passages that belong to a question. For the approach where BiLSTM-CRF layer is used as the Sequence Tagging Layer, we choose answer candidates by using the Viterbi algorithm, and rank candidates based on the probability scores of the LSTM network.

### 5 Experiments

In this section, we first elaborate on our experimental settings. For the rest of the section, we report our experimental results and comparisons between the models with different settings.

**Table 1.** Statistics of list-type questions in the original BioASQ datasets and two different version of pre-processed datasets

| Dataset | Config. | Train | | Test | |
|---------|---------|-------|--------|----------|--------|
|         |         | Question | Sample | Question | Sample |
| BioASQ 7b | Original | 556 | 5324 | 88 | 393 |
|           | Single-span | 529 | 7722 | 88 | 393 |
|           | Seq-Tag | 527 | 3610 | 88 | 393 |
| BioASQ 8b | Original | 644 | 5717 | 75 | 383 |
|           | Single-span | 614 | 8416 | 75 | 383 |
|           | Seq-Tag | 610 | 3914 | 75 | 383 |

*Note*: The column *Sample* denotes the number of data points that are composed of a question and passage pair.

#### 5.1 Datasets

The BioASQ datasets are directly derived from the BioASQ Challenge which is an annual competition for biomedical semantic indexing and QA (Tsatsaronis et al., 2015). Every year, test examples for the BioASQ Challenge are created by biomedical experts and made available to the public as a form of a testing dataset after minor revisions (Nentidis et al., 2020a, b). The training dataset is incrementally built by combining the training and testing datasets of previous versions. For example, the BioASQ 7b training dataset encompasses the training and testing datasets of BioASQ 6b. We used the training and testing sets of BioASQ 6b for hyperparameter searching, and further used those hyperparameters on our experiments conducted on BioASQ 7b. BioASQ 8b experiments are carried out in an analogous fashion.

A single sample contained in the BioASQ datasets contains the question and URL of relevant articles and snippets (http://participants-area.bioasq.org/general_information/Task7b/). Following Lee et al. (2020); Yoon et al. (2019b); Jeong et al. (2020), we retrieved the PubMed articles using provided URLs and used a pair of title and abstract as a passage.

Statistics of list-type questions and data points for the BioASQ datasets are described in Table 1. We formulated a question-passage-answer triplet $(Q, P, A)$ as a sample for training and testing. A triplet contains only one item for a question and a passage element. The number of items for an answer element in a triplet is only one for single-span extraction configuration, while for a sequence-tagging configuration, an answer element can contain multiple items. Specifically, for a list-type question $Q$ with a set of relevant passages $\{P_1, \ldots, P_i\}$, and a list of answer phrases $\{A_1, \ldots, A_j\}$, the number of training samples (i.e. data points) for a question is $i * j$ for a single-

**Table 2.** Performance comparison among the models on the BioASQ 7b and 8b list question datasets

| List-type question | | BioASQ 7b | | | BioASQ 8b | | |
|---|---|---|---|---|---|---|---|
| Language model | System | Prec. | Recall | F1 | Prec. | Recall | F1 |
| BioBERT (Lee *et al.*, 2020) | Yoon *et al.* (2019b) | 0.5941 (0.0072) | 0.3869 (0.0069) | 0.4295 (0.0069) | 0.4476 (0.0186) | 0.3275 (0.0101) | 0.3382 (0.0115) |
| | Jeong *et al.* (2020) | 0.5911 (0.0181) | 0.3966 (0.0074) | 0.4364 (0.0093) | 0.4581 (0.0071) | 0.3335 (0.0049) | 0.3428 (0.0054) |
| | Ours (Seq-Tag Linear) | 0.4247 (0.0112) | 0.5772 (0.0125) | 0.4498 (0.0116) | 0.3888 (0.0105) | 0.5936 (0.0126) | 0.4355 (0.0083) |
| BlueBERT (Peng *et al.*, 2019) | Yoon *et al.* (2019b) | 0.5408 (0.0107) | 0.3668 (0.0050) | 0.4031 (0.0065) | 0.4941 (0.0077) | 0.3335 (0.0053) | 0.3656 (0.0057) |
| | Ours (Seq-Tag Linear) | 0.4048 (0.0064) | 0.6171 (0.0072) | 0.4538 (0.0047) | 0.3368 (0.0089) | 0.5698 (0.0066) | 0.3917 (0.0068) |
| PubMedBERT (Gu *et al.*, 2021) | Yoon *et al.* (2019b) | 0.5709 (0.0099) | 0.3964 (0.0070) | 0.4328 (0.0067) | 0.4754 (0.0115) | 0.3502 (0.0055) | 0.3622 (0.0070) |
| | Ours (Seq-Tag Linear) | 0.4260 (0.0099) | 0.6276 (0.0102) | 0.4758 (0.0088) | 0.3775 (0.0122) | 0.5855 (0.0077) | 0.4254 (0.0085) |

*Note*: Reported scores were micro-averaged across the 10 testing batches. Standard deviations are denoted in the parenthesis. F1 score is the official metric for the list questions of the BioASQ dataset. Note that we used full abstracts as input passages for all systems.

span extraction configuration, and *i* for a sequence-tagging configuration. For the training dataset, we excluded passages from which we could not find any matching answers. We defer details regarding preprocessing to the Supplementary Appendix.

The official evaluation metrics for BioASQ datasets differ according to the question type (http://participants-area.bioasq.org/Tasks/b/eval_meas_2018; https://github.com/BioASQ/Evaluation-Measures). Each example of list-type questions is evaluated based on the precision, recall and F-1 measures. The scores are averaged over the list-type questions and reported as the final score for the system.

### 5.2 Sequential transfer learning

Following Lee *et al.* (2020) and Yoon *et al.* (2019b), we employed sequential transfer learning by first initializing the model's weights using BioBERT weights that were pre-trained on SQuAD v1 dataset (Rajpurkar *et al.*, 2016). We subsequently fine-tuned the pre-trained model on the BioASQ datasets using a smaller learning rate. We expect that this sequential transfer learning strategy will alleviate the data scarcity of the BioASQ datasets.

### 5.3 Experimental details

We used `BioBERT v1.1`, `bluebert_pubmed_uncased_L-12` and `PubMedBERT-base-uncased-abstract` where each has identical parameter settings with the BERT$_{BASE}$ model, as the contextualized representation of our model. Therefore, we have approximately 108 million trainable parameters. We set 512 tokens as the maximum sequence length, and sequences with the length after tokenization exceeds 512 were truncated. A training batch size of 18 was selected. Training samples were randomly shuffled at the start of each epoch. A learning rate of 5e-5 was selected for training SQuAD dataset and 5e-6 was selected for the BioASQ datasets. Maximum training steps were 80 000 for BioASQ7b and 84 000 for BioASQ8b respectively (approximately 400 epochs). We used a single NVIDIA Titan RTX (24GB) GPU for fine-tuning (single run) and the training process took less than 24 h for an experiment.

### 5.4 Results

In this section, we report the experimental results and verify that our approach to reformulate QA as sequence tagging improves the performance of existing QA models on list-type questions.

Table 2 shows the experimental results of our approach and the baselines. In order to minimize the randomness added by the initialization step, we conducted 10 independent runs with identical hyperparameters but with different random seeds, and reported average performances across the runs and the standard deviations of them.

#### 5.4.1 Baselines

Our baseline model from Yoon *et al.* (2019b) is a challenge-winning model (BioASQ 7b) and can be described as an expanded version of the BioBERT model that can answer list-type questions by adding additional post-processing steps to decide the number of answers: a thresholding approach; and a rule-based *number of answers detection* from question strings for such questions containing a number. On top of the previous model, Jeong *et al.* (2020) further added an additional transfer learning step using natural language inference (NLI) datasets and achieved performance improvement over the preceding models.

The original models of Yoon *et al.* (2019b) and Jeong *et al.* (2020) (full abstract model) use the same neural network architecture. They both essentially add a linear token-level classification layer on top of a BioBERT backbone model. They differ in how post-processing and training were conducted. In addition to BioBERT version models, we have experimented with the model from Yoon *et al.* (2019b) using BlueBERT (Peng *et al.*, 2019) and PubMedBERT (Peng *et al.*, 2019) as a backbone model and reported the results.

**Table 3.** Performance of the sequence tagging approach with different sequence tagging layer on BioASQ 8b task

| Tagging Layer | BioASQ 8b List questions | | |
| --- | --- | --- | --- |
| | Precision | Recall | F1 score |
| Linear | 0.3984 (0.0051) | 0.6016 (0.0037) | 0.4402 (0.0047) |
| BiLSTM | 0.4015 (0.0146) | 0.5787 (0.0231) | 0.4370 (0.0121) |
| BiLSTM-CRF | 0.3868 (0.0126) | 0.5925 (0.0072) | 0.4312 (0.0084) |

*Note*: Averages and standard deviations of five independent runs are reported in the table. Standard deviations are denoted in the parenthesis.

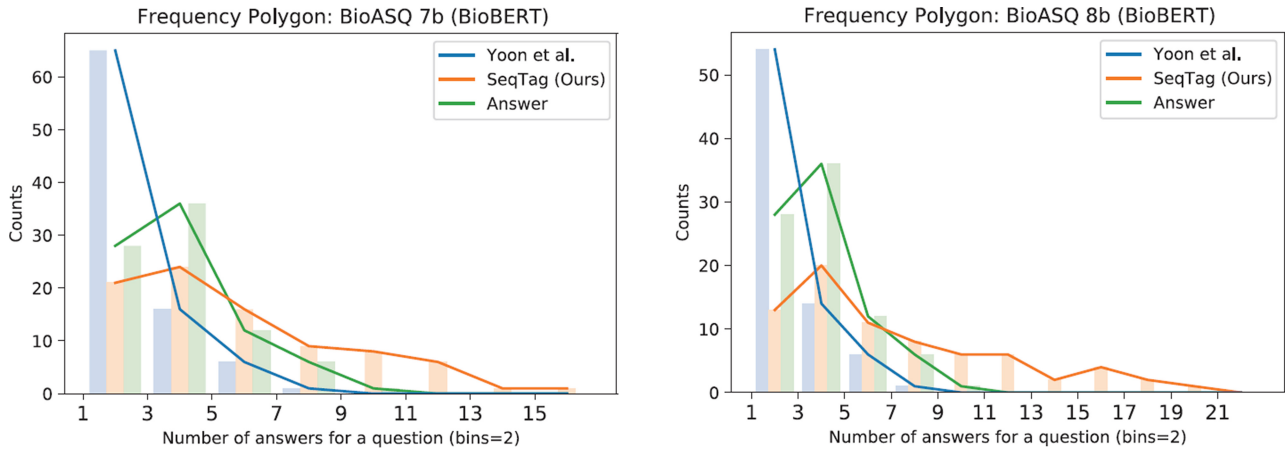

**Fig. 4.** Frequency polygon and histograms of the number of answers (predicted and golden). The predicted answer number distributions of the model predictions are marked as Yoon *et al.* (2019b) and SeqTag (Ours). The answer number distributions of the gold standard (testing dataset) are marked as Answer. The size of a bin is 2.0 and the y-axis indicates the number of questions in the given bin. The number of predicted answers of the baseline model shows the highest population at the first bin [1, 3) whereas our model and the golden answers show the highest population at the second bin [3,5). Questions predicted as having 0 answers are excluded in this graph

#### 5.4.2 Sequence tagging approach

Our experimental results using the fully connected linear layer as Sequence Tagging Layers are displayed in Table 2. Without any assistance from rule-based post-processing steps or complicated three-step transfer learning, our models displayed strong improvement on answering list-type questions. When compared to the model of Yoon *et al.* (2019b), our model using BioBERT, BlueBERT and PubMedBERT as underlying LM achieved performance gains of 2.03%, 5.07% and 4.30% on list questions in BioASQ 7b and 9.73%, 2.61% and 6.32% on BioASQ 8b, respectively (absolute improvement).

### 5.5 Sequence tagging layer

Previous research papers have shown that the application of Recurrent Neural Networks (RNN), such as LSTM, and CRF layer can boost the accuracy of popular sequence tagging tasks, such as NER or part-of-speech tagging (Habibi *et al.*, 2017; Huang *et al.*, 2015; Yoon *et al.*, 2019a). Since our model uses a modified structure of sequence tagging tasks, we explored the effects of RNN-based approaches. Our experimental results using linear, BiLSTM and BiLSTM-CRF models as Sequence Tagging Layers are displayed in Table 3. Following our main experiments, we conducted five independent runs and reported statistics of them. Sequential transfer learning is applied and the model is initiated from SQuAD v1 trained LMs. Special token exclusion (Section 4.2.2) is not applied to these experiments due to CRF layers (For the same reason, we used TensorFlow v1.5 framework for experiments in this section. The rest of the experiments are implemented with pytorch v1.7.).

Our experimental results present that adding LSTM or LSTM-CRF layer does not show significant performance gain over using simple linear layer as output layer. This suggests that solving QA tasks with sequence tagging may have different nature from traditional sequence tagging tasks in the NLP fields. RNN cells encode an input token sequence recurrently. When encoding a token, RNN cells store information of previous input tokens as a fixed-size vector and use the vector in encoding the next token. As suggested by Seo *et al.* (2016), attention flows between question tokens and context tokens are important for solving EQA tasks. Since RNN encodes tokens that compose a question, into a fixed-size vector, the model using RNN/RNN-CRF layer loses the benefits of the transformer structure, where the attentions of each token are connected and share information between them.

## 6 Discussion

In this section, we elaborate on the experimental results from the previous section and analyze the strength of our approach.

### 6.1 Results analysis

The experimental results in Table 2 show that our approach constantly outperforms the previous approach in the Recall metric. These observations correspond to one of the strengths of the proposed model, that the model does not need to rely on thresholding when deciding the answer candidates. For a model using thresholding (threshold = $t$) on the probability, the number of answers is limited to a maximum of $1/t$. On the other hand, our model can predict answers as many as the model needs, lifting the unnatural regulations on the number of questions.

Figure 4 shows the distributions of the number of predicted answers of baseline (Yoon *et al.*, 2019b) and SeqTag model on BioASQ list datasets. As stated in the previous paragraph, baseline models are restricted to predict less than a certain number of answers, resulting in a tendency to predict a smaller number of answers. As a result, the distributions of baseline were inconsistent with the distribution of the testing dataset (denoted as *Answer*). The number of answers is most frequent in the first bin ([1, 2] section) for the baseline, but in our model and the testing dataset, the second bin ([3, 4] section) was the most frequent, providing circumstantial evidence that our model can learn the proper number of answers for a given question.

### 6.2 Limitations of previous works

In addition to the thresholding, Yoon *et al.* (2019b) and Jeong *et al.* (2020) utilized the rule-based post-processing steps, namely *number of answers detection*, to handle list-type questions. For example, in order to answer the question 'What **2** biological processes are regulated by STAMP2 in adipocytes?', the model detects any number in the question, two (**2**), and outputs answers as many as the detected number. However, models with number of answers detection are susceptible to the proportion of questions with answers that have numbers. For the BioASQ 8b testing dataset, the proportion is nearly half than that of BioASQ 7b (Table 4). We credit the extra performance gains of our model on BioASQ 8b to not using rule-based post-processing steps and the robustness attained from it.

The testing dataset for the BioASQ challenge provides both document identifiers, namely PMID's, and snippets which are manually curated sentences within the provided document. Participants of the BioASQ challenge can choose which data to use. Jeong *et al.* (2020) won the BioASQ 8b challenge by using snippets as the source of the passage and exhibited significant performance gains. However, in this article, we do not consider snippets as the passage since the use of snippets may not sufficiently fulfill the final goal of EQA to build an automated process of finding an answer from the given document. (We defer more details regarding the snippet approach to the Supplementary Appendix.)

### 6.3 Universal modeling for extractive QA

Employing sequence tagging for EQA enables a single unified model to answer factoid- and list-type questions without any task-specific layer modification. In order to experiment on the usability of our model on the universal setting, we simply amalgamated two datasets and trained our model on the merged dataset. More specifically, our setting is similar to multi-task learning in the sense that we use various data sources, but does not require task-specific layers.

Table 5 shows the performance and the utility of sequence tagging models trained on different training dataset combinations. Model of Experiment (1) is a standard sequence tagging model trained on the BioASQ List 8b dataset (Identical to the Seq-Tag BioBERT model in Table 2). Likewise, model of (2) is trained on the BioASQ Factoid 8b dataset. Model of Experiment (3) has an identical structure with Model of (1) and (2) but is trained on the amalgamated dataset (List + Factoid). Without any further fine-tuning steps, Model (3) showed competitive performance with model (1) on list questions, suggesting that our modeling can be used for a universal approach for both factoid and list questions (i.e. all questions regardless of the expected number of answers). Our question-type agnostic approach provides additional benefits for applications of QA models to real-world use, as metadata (or prior knowledge) of naturally posed biomedical questions are not given in general.

## 7 Conclusion

In this article, we proposed a sequence tagging approach for BioEQA. In our preliminary study, we displayed that list-type questions, which are questions with multiple phrases as an answer set, are more abundant in the biomedical context than questions in the general domain, and take up a significant portion of questions. Stemming out from the preliminary study, we stress that solving list-type questions are a key building block for modeling robust BioEQA systems.

We proposed and demonstrated the advantage of the sequence tagging approach in predicting a variable number of phrases as answers for a question. Our proposed approach outperformed baseline models in a large gap, regardless of the backbone models (BioBERT, BlueBERT and PubMedBERT). Average performance improvements over baseline model (Yoon *et al.*, 2019b) were 3.80% for BioASQ 7b List questions and 6.22% for BioASQ 8b List questions (F1 score). Moreover, the sequence tagging approach enables models to handle both list and factoid questions using a unified structure, suggesting that list questions can be viewed as a generalized version of factoid questions.

### Acknowledgements

The authors express gratitude toward Sean S. Yi and Hyunjae Kim for detailed feedback of the manuscript.

### Data availability

The data underlying this article are available at https://github.com/dmis-lab/SeqTagQA.

### Funding

This work was supported by the National Research Foundation of Korea [NRF-2020R1A2C3010638, NRF-2014M3C9A3063541]; the Korea Health Technology R&D Project [HR20C0021] through the Korea Health Industry Development Institute, funded by the Ministry of Health & Welfare, Republic of Korea; and the Research Collaboration Project from AstraZeneca UK.

*Conflict of Interest*: W.Y. has received research funding (Studentship) from AstraZeneca UK.

**Table 4.** The proportion of list-type questions with the number of answers in the question, out of list questions in testing datasets

| Dataset | Test question | Question with number | Proportion |
| --- | --- | --- | --- |
| BioASQ 7b | 88 | 23 | 26.1% |
| BioASQ 8b | 75 | 10 | 13.3% |

**Table 5.** Performance of our model on multiple data

| | | BioASQ 8b test | | |
| --- | --- | --- | --- | --- |
| | Training data | List-F1 | Factoid-MRR | Utility |
| (1) | List 8b | 0.4310 (0.0056) | — | Focused on multi-answer questions (Requires metadata) |
| (2) | Factoid 8b | — | 0.3759 (0.0034) | Focused on single-answer questions (Requires metadata) |
| (3) | List 8b + Factoid 8b | 0.4148 (0.0081) | 0.3795 (0.0183) | **Universally usable** regardless of expected number of answers |
| | Difference | −0.0162 | 0.0036 | |

*Note*: Our sequence tagging approach enables to train a universal model that can predict questions without knowing metadata on the given question. Statistics on five individual runs are reported in the table.